\documentclass[runningheads]{llncs}
\usepackage{enumitem}
\usepackage{amsmath}
\usepackage{graphicx}
\usepackage[caption=false]{subfig}
\usepackage{arydshln}
\usepackage{array}
\newcolumntype{P}[1]{>{\centering\arraybackslash}p{#1}}

\begin{document}
\title{Fast Distance-based Anomaly Detection in Images Using an Inception-like Autoencoder}
\titlerunning{Fast Distance-based Anomaly Detection in Images}
%
\author{Natasa Sarafijanovic-Djukic\inst{1}
\and
Jesse Davis\inst{2}\orcidID{0000-0002-3748-9263} }
\authorrunning{Sarafijanovic-Djukic and Davis}
%
\institute{IRIS Technology Solutions, Barcelona, Spain \\
\email{natasa.sdj@iris.cat} \and
KU Leuven, Belgium \\
\email{jesse.davis@cs.kuleuven.be}
}
\maketitle              
\begin{abstract}
The goal of anomaly detection is to identify examples that deviate from normal or expected behavior. We tackle this problem for images. We consider a two-phase approach. 
First, using normal examples, a convolutional autoencoder (CAE) is trained to extract a low-dimensional representation of the images. Here, we propose a novel architectural choice when designing the CAE, an Inception-like CAE. It combines convolutional filters of different kernel sizes and it uses a Global Average Pooling (GAP) operation to extract the representations from the CAE's bottleneck layer.
Second, we employ a distanced-based anomaly detector in the low-dimensional space of the learned representation for the images. 
However, instead of computing the exact distance, we compute an approximate distance using product quantization. This alleviates the high memory and prediction time costs of distance-based anomaly detectors. 
We compare our proposed approach to a number of baselines and state-of-the-art methods on four image datasets, and we find that our approach resulted in improved predictive performance.

\end{abstract}
\section{Introduction}
\label{introduction}
The goal of anomaly detection~\cite{chandola2009anomaly,DBLP:conf/icdm/VercruyssenMVMB18} is to identify examples that deviate from what is normal or expected. We tackle this problem for images which is relevant for applications such as visual quality inspection in manufacturing \cite{haselmann2018anomaly}, surveillance \cite{shashikar2017traffic,sultani2018real}, biomedical applications \cite{taboada2009anomaly,wei2018anomaly}, self-driving cars \cite{creusot2015real}, or robots \cite{chakravarty2007anomaly,munawar2017spatio}. This has motivated significant interest in this problem in recent years. 


The classic approach to anomaly detection is to treat it as an unsupervised problem (e.g.,~\cite{breunig2000lof,ramaswamy2000efficient}) or one-class problem~\cite{ruff2018deep,Tax:2004:SVD:960091.960109}.  
Recently, there has been a surge of interest in applying deep learning to anomaly detection, particularly in the context of images (e.g.,~\cite{Golan2018DeepAD,ruff2018deep,sabokrou2016fully,seebock2016identifying}).  
In this line of work, one strategy is to use (convolutional) autoencoders, which is typically done in one of two ways. First, it is possible to directly use the autoencoder as an anomaly detector. This can be done by using an example's reconstruction error as the anomaly score (e.g.,~\cite{xia2015learning}). Second, the autoencoder can be used to learn a new low-dimensional representation of the data after which a classical anomaly detection approach is applied on top of this learned representation (e.g.,~\cite{andrews2016detecting,xu2015learning}). 

In this paper, we take the one-class approach and follow the second strategy for the problem of detecting anomalous images. We begin by training a convolutional autoencoder (CAE) on only normal images. Here our contribution is to propose a novel CAE architecture. It is inspired by the Inception classification model~\cite{szegedy2015going} that combines convolutional filters of different kernel sizes. Once the CAE is trained, we use a Global Average Pooling (GAP) operation to extract the low-dimensional representation from the CAE's bottleneck layer. In contrast, existing approaches directly use the bottleneck layer's output. Using the GAP operation is motivated by its successes in reducing overfitting in classification CNN models~\cite{lin2013network}, and  extracting image representations from the hidden layers of pretrained classification models for image captioning~\cite{vinyals2015show}. 

At test time, we use a classic nearest-neighbor distanced-based anomaly detector~\cite{ramaswamy2000efficient} in the learned low-dimensional representation space. Here our contribution is to compute an approximate distance using product quantization~\cite{DBLP:journals/pami/JegouDS11}, which improves the runtime performance and memory footprint 
of this approach compared to using the exact distance. Empirically, we compare our proposed approach to a number of existing approaches on four standard datasets used for benchmarking anomaly detection models for images. We find that our approach generally achieves better predictive performance. 

\section{Background and Related Work}
\label{relwork}
This work draws on ideas from anomaly detection both in general and for images, deep learning, and fast nearest neighbors. We now review each of these areas. 

\subsection{Anomaly Detection}
For a variety of reasons (e.g., what is anomalous changes over time or expense), it is often difficult to obtain labels for examples belonging to the anomaly class. Therefore, anomaly detection is often approached from an unsupervised~~\cite{breunig2000lof,ramaswamy2000efficient} or one-class perspective~\cite{Tax:2004:SVD:960091.960109}

In order to identify anomalies, unsupervised approaches typically assume that anomalous examples are rarer and different in some respect than normal examples. 
One standard approach is to assume that anomalies are far away from normal examples or that they lie in a low-density region of the instance-space~\cite{breunig2000lof,ramaswamy2000efficient}. A common approach~\cite{ramaswamy2000efficient} that uses this intuition  is based on  $k$-nearest neighbors. This algorithm produces a ranking of how anomalous each example is by computing an example's distance to its  $k^{\textrm{th}}$ nearest neighbor in the data set.  Despite its simplicity, this approach seems to work very well empirically~\cite{campos2016evaluation}.

The idea underlying one-class-based anomaly detection is that the training data only contains normal examples. Under this assumption, the training phase attempts to learn a model of what constitutes normal behavior. Then, at test time, examples that do not conform to the model of normal are considered to be anomalous. One way to do this is to use a one-class SVM~\cite{Tax:2004:SVD:960091.960109}. 

\subsection{Deep Learning for Anomaly Detection}
\label{subsec:deepLearningAD}
Autoencoders are the most prevalent deep learning methods used for anomaly detection. An autoencoder (AE) is a multi-layer neural network that is trained such that the output layer is able to reproduce its input. An AE has a bottleneck layer with a lower dimension than the input layer and hence allows learning a low-dimensional representation (encoding) of the input data. 

AEs are used for anomaly detection in images in two ways. First, an AE can be directly used as an anomaly detector. Here, a typical way to assign an anomaly score for a test example is to apply the AE and calculate the example's reconstruction error (e.g., the mean squared error between the example and the AE's  output~\cite{sakurada2014anomaly,xia2015learning}). 
Second, an AE can be used as part of a two-step process: 1) train an AE on the training data; and 2) learn a standard (shallow) anomaly detector on the transformed training data~\cite{andrews2016detecting,xu2015learning}. We follow this strategy.

Deep approaches to anomaly detection for image data often  use a convolutional autoencoder (CAE) which include convolutional layers in the AE architecture~\cite{richter2017safe,seebock2016identifying}.  Another line of work uses Generative Adversarial Networks (GAN) for this task~\cite{deecke2018anomaly,sabokrou2018adversarially,schlegl2017unsupervised}. This two-step process is also used to make the density estimation task easier by learning low-dimensional representations.  

Recently, there have been attempts to design fully end-to-end deep models for anomaly detection. Deep  Support  Vector  Data Description (Deep SVDD) \cite{ruff2018deep} is trained using an anomaly detection based objective that minimizes the volume of a hypersphere enclosing the data representations. Deep Autoencoding Gaussian Mixture Model (DAGMM)~\cite{zong2018deep} uses the representation layer of a deep autoencoder in order to estimate the parameters of a Gaussian mixture model, by jointly optimizing parameters of the autoencoder and the mixture model.  

Deep Structured Energy Based Model (DSEBM)~\cite{zhai2016dsebm} belongs to the group of the energy-based models, a powerful tool for density estimation. The energy-based models make a specific parameterization of the negative log probability, which is called the energy, and then compute the density with a proper normalization. In DSEBM, the  energy  function  is   a  deep  neural network.

Another method \cite{Golan2018DeepAD} uses a data augmentation, and generates new training examples by applying a number of geometric transformations to each training example. Then, a multi-class neural network is trained to discriminate among the original images, and all of the geometric transformations applied to the images. Given a test image, the same transformations are applied to it and the prediction is made based on the network's softmax activation statistics. 

\subsection{Fast Nearest Neighbors Search}
A notable potential issue with nearest-neighbors-based approaches is that finding the nearest neighbor at test time can be very computationally expensive, particularly for large training sets or high-dimensional examples. Consequently, there has been substantial interest in developing efficient approaches for performing this search either exactly (e.g., using a kd-tree or other index structure) or approximately~\cite{DBLP:conf/vldb/GionisIM99,DBLP:journals/pami/JegouDS11}. 

One prominent recent approach is product quantization~\cite{DBLP:journals/pami/JegouDS11}. This approach works by compressing the training data by partitioning the features used to describe the training example into $m$ equal width groups. It then learns a code book for each partition. Typically, this is done by running k-means clustering on each partition which only considers the features assigned to that partition. Then the values of all features in the partition are replaced by a single $c$-bit code representing the cluster id that the example is assigned to in the current partition. Hence each example is rerepresented by $m$ $c$-bit code words. 

At test time, finding a test example's nearest neighbor using the (squared) L2 distance can be done efficiently by using table look-ups and addition. For a test example, a look up table is constructed for each partition that stores the squared L2 distance to each of the $k = 2^c$ cluster centroids in that partition. Then the approximate distance to each training example is computed using these look up tables and the nearest example is returned. 

Locality-sensitive hashing improves efficiency by using hashing to identify a limited number of likely candidate nearest neighbors. Then, a test example is only compared to those examples. Hence, the approximation comes from the fact that not all training examples are considered as the possible nearest neighbor. People have investigated incorporating hashing-based techniques into distance-based anomaly detection systems~\cite{Hachiya2013NSHNS,schubert:dsaa15,zhanga:pr2016}.

\section{Our Approach}
\label{model}
At a high-level, our approach has two steps: extracting a low dimensional image representation and assigning a distance based anomaly score.

\paragraph{Extracting a low-dimensional image representation.} Given a training set of normal image examples, an Inception-like convolutional autoencoder (InceptionCAE) is trained that minimizes the mean squared reconstruction error on the training data. Once the InceptionCAE is trained, the GAP operation is applied on its bottleneck layer to extract a  low-dimensional image representation vector.  
\paragraph{Assigning a distance-based anomaly score.} 
An distance-based anomaly score is assigned using the learned representation vectors for the images.
First, the trained InceptionCAE model is used to convert all training images to the learned low-dimensional representation. Second, it converts a given test image into the same space and assigns an anomaly score by computing the quantized Euclidean distance between the test image and its nearest neighbor in this space. 


\subsection{Inception-like Convolutional Autoencoder}
\label{sec:inceptionCae}

When using a CAE in a two-step anomaly detection approach, the detector's predictive performance clearly depends on the quality of the learned low-dimensional representation.
In supervised image classification, sophisticated deep architectures such as Inception (GoogleNet) \cite{szegedy2015going}, Residual Networks \cite{he2016deep} or DenseNet \cite{huang2017densely}, have yielded considerable performance gains over a basic CNN architecture.
Hence, we expect that adapting these techniques to the CAE setting could improve the quality of the CAE's learned low-dimensional representation. 

\begin{figure}[h!]
\begin{center}
\includegraphics[width = 0.6 \linewidth]{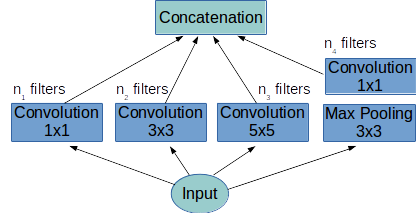}
\caption{Inception-like layer.}
\label{fig:inception}
\end{center}
\end{figure}

Inspired by the Inception architecture, we design an Inception-like CAE architecture that combines convolutional filters of different kernel sizes. The main unit of this architecture is an Inception-like layer shown in Figure \ref{fig:inception}, where it combines outputs from $1\times1$, $3\times3$ and $5\times5$ convolutions as well as a maximum pooling operation. 

Table \ref{tab:cae-gap-incept} outlines the details of our Inception-like CAE architecture. We make our architecture as similar as possible to the baseline CAE architecture \cite{ruff2018deep} and it has the same number of layers and the same number of convolutional filters in each layer to enable a fair comparison.  
Here, \texttt{Inception(n)} denotes the Inception-like layer with $n_1=n_2=n_3=n_4=n$. 
Each convolution operation is followed by Batch Normalization and a Leaky ReLU activation, except the last layer which has a Sigmoid activation. 

Beside the architectural change of introducing the Inception-like layer into a CAE, another subtlety in our approach is how we extract the low-dimensional image representation from the CAE. Existing approaches extract a learned image representation simply by using the output of the CAE's bottleneck layer. Consequently, the CAE architecture must be designed such that the bottleneck layer matches the desired dimension of the learned image representation. Our approach extracts the learned image representation by applying a Global Average Pooling (GAP) operation to the output of the CAE's bottleneck layer. The GAP operation on a tensor of the dimension $a\times b\times c$ results in a  vector of the dimension $1\times c$, where each component is an average value over  the tensor slice of the dimension $a \times b$ that corresponds to this component. Hence, using the GAP operation as an extractor permits using a wider bottleneck layer than existing CAE's architectures do.  

Our intuition behind this architectural choice is that having a wider bottleneck will permit retaining some information that a narrower bottleneck would filter. Thus, the GAP operation on this wider bottleneck would yield a better learned representation. Though the use of GAP is not a novel in deep learning architectures, our contribution is to study its use in conjunction with an CAE for extracting image representations.

 The Inception-like CAE is trained on the normal training images, where the objective is to minimize the mean squared error between the input and the output. Applying the GAP operation on the trained network's bottleneck yields a 128-dimensional learned image representation. Note that the GAP operation allows having a wider bottleneck layer in our Inception-linke CAE architecture than in the baseline CAE architecture ($4\times4\times128$ versus $1\times128$). 

Our experiments show that each of our two architectural choices in designing our CAE contribute to improved anomaly detection  performance.

 \begin{table}
\begin{center}
\begin{tabular}{| p{5cm} | p{3cm} |}  

\cline{1-2} 
InceptionCAE & Output dimension \\ 
\cline{1-2} 
Input Layer & $32\times32\times n_{\text{channels}}$ \\
Inception (8). MaxPooling(2,2). & $16\times16\times32$ \\
Inception(16). MaxPooling(2,2). & $8\times8\times64$  \\
Inception(32). MaxPooling(2,2). & $4\times4\times128$ \\
\cline{1-2} 
Inception(32). Upsampling(2,2). & $8\times8\times128$ \\
Inception(16). Upsampling(2,2).& $16\times16\times64$ \\
Inception (8). Upsampling(2,2). & $32\times32\times32$ \\ 
Convolution2D($n_{channels}$) & $32\times32\times n_{\text{channels}}$ \\
\cline{1-2} 
\end{tabular}
\end{center}
\caption {Inception-like CAE.}  \label{tab:cae-gap-incept}
\end{table}
\subsection{Approximated Distance-Based Anomaly Detection}

We assign an anomaly score to a test example by operating on the extracted images representations and not on the raw data itself.  Specifically, a test example's anomaly score is the quantized (squared) Euclidean distance in the learned representation space to its nearest neighbor in the training data. The primary advantage of using product quantization instead of the exact distance is that it is substantially faster to compute (at the expense of being an approximation). 

Hashing-based solutions have been extensively explored to speed-up distance-based neighbor approaches. While extensively used in nearest-neighbor search, product quantization has received little attention within anomaly detection. One advantage of quantization over a hashing based solution is that it still compares a test example to each training example, it just does so in an approximate manner. 

Quantization may provide another benefit when the training data only contains normal examples: its approximation may enforce some regularization on the training data. That is, by mapping each partition of the example to a prototype it may smooth out some variation and make the examples look more "normal," which is beneficial if true. In some cases, we observe empirically that the quantization does indeed improve performance. 

\section{Experimental Results}
\label{results}

Our empirical evaluation addresses the following questions:\\
\textbf{Q1} How does our proposed approach compare to existing anomaly detection techniques for images?\\ 
\textbf{Q2} What is the effect of using product quantization to approximate the distance calculation on performance? 
\\
\textbf{Q3} What is the effect of using the distance-based nearest neighbors approach to assign an anomaly score compared to using an example's reconstruction error? \\
\textbf{Q4} How sensitive is the performance of our approach to changes in the quantization parameters? 

To address these questions, we compare our proposed approach to a number of shallow and deep baselines on four standard benchmark datasets. Next, we describe the approaches, data, methodology and results in greater detail.

\subsection{Methods Compared} \label{subsec:methods}
 
The main empirical comparison considers the following methods. \\
\textbf{Raw NN-QED}: This shallow approach corresponds to applying the classic kNN-based anomaly detection~\cite{ramaswamy2000efficient} on the raw image data, except that it uses an approximate distance measure. It assigns an anomaly score to a test example as the quantized squared Euclidean distance (QED) to the test example's nearest neighbor in the raw training images.\\
\textbf{DeepSVDD}: This method is a deep extension of the support vector data description method \cite{ruff2018deep}. We use the same baseline CAE architecture for all the datasets as the one used for a CIFAR-10 dataset in the respective paper.\\
\textbf{DSEBM}: This method is a deep extension of energy based models \cite{zhai2016dsebm}, where  we adjust a neural network to correspond to the baseline CAE architecture used in Deep SVDD in order to have a fair comparison.\\
\textbf{CAE OCSVM}: This method trains an CAE with the same baseline architecture as in DeepSVDD. Then the learned image data representations obtained from the output of the CAE's bottleneck layer are used as the input to OCSVM. \\
\textbf{CAE NN-QED}: This method trains an CAE with the same baseline architecture as in DeepSVDD. A test example's anomaly score is calculated as the quantized squared Euclidean distance in the CAE's learned representation space to its nearest neighbor in the training set.\\
\textbf{InceptionCAE NN-QED}: This is our approach. It uses our proposed InceptionCAE architecture outlined in Section~\ref{sec:inceptionCae}. A test example's anomaly score is calculated as the quantized squared Euclidean distance in the InceptionCAE's learned representation space to its nearest neighbor in the training set.\\

\subsection{Datasets}

Our experiments use four common benchmark datasets for both deep learning and anomaly detection approaches.  MNIST~\cite{lecun2010mnist} and Fashion~MNIST~\cite{xiao2017fashion} contains ten classes and have fixed train-test splits with the training set containing 60,000 examples (6,000 examples for each class) and the test set 10,000 examples (1,000 for each class). CIFAR10 has ten class, while in CIFAR100 we consider the 20 super-classes~\cite{krizhevsky2009learning}. Both have fixed train-test splits with the training set containing 50,000 examples and the test set 10,000 examples. All the datasets are completely labeled which enables computing standard evaluation metrics. 

\begin{table}
\begin{center}
\begin{tabular}{ | P{1.3cm} | P{1.5cm} || P{1.8cm} | P{1.8cm} | P{1.8cm} || P{2.2cm} |}
\cline{1-6} 

Dataset & Normal Class & DSEBM & CAE OCSVM & DeepSVDD & InceptionCAE NN-QED \\ \cline{1-6}

& 0    & 94.9 $\pm$ 4.0  & 95.4 $\pm$ 0.8 & 99.1 $\pm$ 0.1  & 98.7 $\pm$ 0.3 \\
& 1    & 98.7 $\pm$ 0.1  & 97.4 $\pm$ 0.3 & 99.7 $\pm$ 0.0  & 99.7 $\pm$ 0.0 \\
& 2    & 69.0 $\pm$ 11.5 & 77.6 $\pm$ 3.3 & 95.4 $\pm$ 0.3  & 96.7 $\pm$ 0.7 \\
& 3    & 80.2 $\pm$ 9.7  & 88.6 $\pm$ 1.6 & 95.1 $\pm$ 0.5  & 95.2 $\pm$ 0.4 \\
& 4    & 83.3 $\pm$ 9.1  & 83.6 $\pm$ 1.8 & 95.9 $\pm$ 0.5  & 95.0 $\pm$ 0.5 \\ MNIST
& 5    & 67.4 $\pm$ 6.8  & 71.3 $\pm$ 1.8 & 92.1 $\pm$ 0.5  & 95.2 $\pm$ 0.5 \\
& 6    & 85.6 $\pm$ 5.9  & 90.1 $\pm$ 1.6 & 98.5 $\pm$ 0.1  & 98.3 $\pm$ 0.2 \\
& 7    & 90.4 $\pm$ 2.1  & 87.2 $\pm$ 0.8 & 96.2 $\pm$ 0.4  & 97.0 $\pm$ 0.3 \\
& 8    & 72.1 $\pm$ 7.3  & 86.5 $\pm$ 1.6 & 95.7 $\pm$ 0.4  & 96.2 $\pm$ 0.2 \\
& 9    & 86.8 $\pm$ 2.9  & 87.3 $\pm$ 1.0 & 97.7 $\pm$ 0.1  & 97.0 $\pm$ 0.2 \\ \cdashline{2-6}
& Average   & 82.8 $\pm$ 12.3 & 86.5 $\pm$ 7.5 & 96.6 $\pm$ 2.1  & 96.9 $\pm$ 1.6 \\ \cline{1-6} 
    
& 0    &  89.2 $\pm$  0.1 & 88.0 $\pm$ 0.4 & 98.8 $\pm$ 0.2  & 92.4 $\pm$ 0.4 \\
& 1    &  97.4 $\pm$  0.1 & 97.3 $\pm$ 0.2 & 99.7 $\pm$ 0.0  & 98.8 $\pm$ 0.1 \\
& 2    &  86.0 $\pm$  0.3 & 85.5 $\pm$ 0.8 & 93.5 $\pm$ 1.4  & 90.0 $\pm$ 0.6 \\
& 3    &  90.5 $\pm$  0.1 & 90.0 $\pm$ 0.5 & 94.9 $\pm$ 0.3  & 95.0 $\pm$ 0.3 \\
& 4    &  88.5 $\pm$  0.3 & 88.5 $\pm$ 0.5 & 95.1 $\pm$ 0.6  & 92.0 $\pm$ 0.4 \\ Fashion
& 5	   &  82.4 $\pm$  9.2 & 87.2 $\pm$ 0.7 & 90.4 $\pm$ 0.8  & 93.4 $\pm$ 0.3 \\ MNIST
& 6    &  77.7 $\pm$  1.5 & 78.8 $\pm$ 0.7 & 98.0 $\pm$ 0.2  & 85.5 $\pm$ 0.4 \\
& 7    &  98.1 $\pm$  0.1 & 97.7 $\pm$ 0.1 & 96.0 $\pm$ 0.2  & 98.6 $\pm$ 0.1 \\
& 8	   &  78.8 $\pm$  6.8 & 85.8 $\pm$ 1.4 & 95.4 $\pm$ 0.4  & 95.1 $\pm$ 0.4 \\
& 9    &  96.0 $\pm$  2.7 & 98.0 $\pm$ 0.2 & 97.6 $\pm$ 0.2  & 97.7 $\pm$ 0.2 \\ \cdashline{2-6}   
& Average  &  88.5 $\pm$ 7.9 & 89.7 $\pm$ 6.0 & 95.9 $\pm$ 3.8  & 93.9 $\pm$ 3.9 \\ \cline{1-6} 

\end{tabular}
\end{center}
\caption {Average AUC-ROC and its standard deviation for state-of-the-art deep baselines and our approach on the MNIST and Fashion MNIST datasets.} \label{tab:auc_mnist_fmnist}
\end{table}

Following past work on anomaly detection~\cite{Golan2018DeepAD,ruff2018deep}, we denote the images of one class as normal, while images for all other classes are considered anomalous. The training phase only uses images from the normal class. At test time, test images of all classes are used.

\subsection{Parameters and Implementations}

For all the CAE architectures, we employ the same training procedure as in Ruff et al.~\cite{ruff2018deep}, with a two-phase learning rate schedule (searching + fine-tuning) with initial learning rate $\nu = 10^{-4}$, and subsequently $\nu = 10^{-5}$. We train 100+50 epochs for MNIST and Fashion MNIST, and 250 + 100 epochs for CIFAR-10 and CIFAR-100. Leaky ReLU activations use a leakiness of $\alpha=0.1$. We use a batch size of 200 and set the weight decay hyperparameter $\lambda = 10^{-6}$, and we use an Adam optimization procedure \cite{kingma2014adam}. For CIFAR-10 and CIFAR-100, both the CAE-GAP and InceptionCAE architectures are trained without the GAP layer, but the GAP operation on the bottleneck layer is used at prediction time to extract the image representation. For MNIST and Fashion MNIST, the GAP layer must be included during training to ensure that the bottleneck layer is narrower than the input layer. We implemented the CAEs in the Keras framework \cite{chollet2015keras}. 


We use the Facebook AI Similarity Search (FAISS) library \cite{JDH17} for computing the quantized Euclidean distance using the parameters $m=32$ and $c=4$. We show the effects of these parameters in Subsection~\ref{sec:quantParams}. 

The OCSVM implementation uses the default parameters of Python \texttt{sklearn} library, with radial basis function kernel with $\gamma = 1/n_{features}$ and $\nu=0.5$. 

 Because we use an identical train-test split, we simply report the AUC-ROCs for prior results for DeepSVDD on CIFAR-10 from the paper. For DeepSVDD, we re-run the experiments for the other datasets using the authors' software in order to use the same CAE baseline architecture. Our code is available online.\footnote{\url{https://github.com/natasasdj/anomalyDetection}}

\begin{table}
\begin{center}
\begin{tabular}{ | P{1.3cm} | P{1.5cm} || P{1.8cm} | P{1.8cm} | P{1.8cm} || P{2.2cm} |}
\cline{1-6} 
Dataset & Normal Class & DSEBM & CAE OCSVM & DeepSVDD & InceptionCAE NN-QED   \\ \cline{1-6} 
& 0     & 64.1 $\pm$ 1.5 & 62.4 $\pm$ 0.9    & 61.7 $\pm$ 4.1 & 66.7 $\pm$ 1.3 \\
& 1     & 50.1 $\pm$ 5.1 & 44.4 $\pm$ 1.0    & 65.9 $\pm$ 2.1 & 71.3 $\pm$ 1.3 \\
& 2     & 61.5 $\pm$ 0.8 & 64.2 $\pm$ 0.3    & 50.8 $\pm$ 0.8 & 66.8 $\pm$ 0.6 \\
& 3     & 51.2 $\pm$ 3.0 & 50.7 $\pm$ 0.8    & 59.1 $\pm$ 1.4 & 64.1 $\pm$ 0.9 \\
& 4     & 73.2 $\pm$ 0.5 & 74.8 $\pm$ 0.2    & 60.9 $\pm$ 1.1 & 72.3 $\pm$ 0.8 \\ CIFAR 
& 5     & 54.6 $\pm$ 2.8 & 50.9 $\pm$ 0.5    & 65.7 $\pm$ 2.5 & 65.3 $\pm$ 0.9 \\ 10
& 6     & 68.2 $\pm$ 1.1 & 72.4 $\pm$ 0.3    & 67.7 $\pm$ 2.6 & 76.4 $\pm$ 0.8 \\
& 7     & 52.8 $\pm$ 1.3 & 51.0 $\pm$ 0.7    & 67.3 $\pm$ 0.9 & 63.7 $\pm$ 0.7 \\
& 8     & 73.7 $\pm$ 1.9 & 67.0 $\pm$ 1.6    & 75.9 $\pm$ 1.2 & 76.9 $\pm$ 0.6 \\
& 9	    & 63.9 $\pm$ 5.9 & 50.8 $\pm$ 2.5    & 73.1 $\pm$ 1.2 & 72.5 $\pm$ 1.0 \\ \cdashline{2-6}
& Average   & 61.3 $\pm$ 8.9 & 58.9 $\pm$ 10.1   & 64.8 $\pm$ 7.2 & 69.6 $\pm$ 4.8 \\ \cline{1-6}

& 0     & 63.8 $\pm$ 0.4 & 63.6 $\pm$ 1.2 & 57.4 $\pm$ 2.4 & 66.0 $\pm$ 1.5 \\
& 1     & 48.4 $\pm$ 0.9 & 51.4 $\pm$ 0.7 & 63.0 $\pm$ 1.2 & 60.1 $\pm$ 1.5 \\
& 2     & 63.6 $\pm$ 7.6 & 54.5 $\pm$ 1.0 & 70.0 $\pm$ 3.2 & 59.2 $\pm$ 3.1 \\
& 3     & 50.4 $\pm$ 3.2 & 48.4 $\pm$ 0.8 & 55.8 $\pm$ 2.5 & 58.7 $\pm$ 0.5 \\
& 4     & 57.3 $\pm$ 9.6 & 49.9 $\pm$ 1.3 & 69.0 $\pm$ 1.9 & 60.9 $\pm$ 1.9 \\ 
& 5     & 44.4 $\pm$ 3.3 & 45.3 $\pm$ 1.4 & 51.0 $\pm$ 2.0 & 54.2 $\pm$ 1.3 \\ 
& 6     & 53.3 $\pm$ 5.2 & 53.1 $\pm$ 1.6 & 59.9 $\pm$ 3.3 & 63.7 $\pm$ 1.4 \\
& 7     & 53.4 $\pm$ 1.3 & 58.8 $\pm$ 0.6 & 53.0 $\pm$ 1.2 & 66.1 $\pm$ 1.3 \\
& 8     & 66.9 $\pm$ 0.3 & 67.8 $\pm$ 0.5 & 51.6 $\pm$ 3.2 & 74.8 $\pm$ 0.4 \\
& 9     & 72.7 $\pm$ 4.0 & 70.1 $\pm$ 1.2 & 72.9 $\pm$ 1.5 & 78.3 $\pm$ 0.7 \\ CIFAR 
& 10    & 76.2 $\pm$ 3.4 & 76.7 $\pm$ 0.6 & 81.5 $\pm$ 1.9 & 80.4 $\pm$ 0.9 \\ 100
& 11    & 62.2 $\pm$ 1.2 & 59.7 $\pm$ 0.6 & 53.6 $\pm$ 0.7 & 68.3 $\pm$ 0.6 \\
& 12    & 66.9 $\pm$ 0.4 & 68.2 $\pm$ 0.3 & 50.6 $\pm$ 1.2 & 75.6 $\pm$ 0.7 \\
& 13    & 53.1 $\pm$ 0.7 & 60.6 $\pm$ 0.4 & 44.0 $\pm$ 1.2 & 61.0 $\pm$ 0.9 \\
& 14    & 44.7 $\pm$ 0.7 & 47.1 $\pm$ 1.1 & 57.2 $\pm$ 1.1 & 64.3 $\pm$ 0.7 \\
& 15    & 56.6 $\pm$ 0.2 & 59.7 $\pm$ 0.3 & 47.7 $\pm$ 0.9 & 66.3 $\pm$ 0.4 \\
& 16    & 63.1 $\pm$ 0.4  & 66.0 $\pm$ 0.4 & 54.3 $\pm$ 0.8 & 72.0 $\pm$ 0.5 \\
& 17    & 73.5 $\pm$ 3.6  & 69.4 $\pm$ 1.1 & 74.7 $\pm$ 2.0 & 75.9 $\pm$ 0.7 \\
& 18    & 55.6 $\pm$ 2.2  & 54.5 $\pm$ 0.8 & 52.1 $\pm$ 1.7 & 67.4 $\pm$ 0.8 \\
& 19    & 57.3 $\pm$ 1.6  & 54.7 $\pm$ 1.0 & 57.9 $\pm$ 1.8 & 65.8 $\pm$ 0.6 \\ \cdashline{2-6}
& Average   & 59.2 $\pm$ 9.3  & 59.0 $\pm$ 8.6 & 58.9 $\pm$ 9.9 & 67.0 $\pm$ 7.1 \\ \cline{1-6} 

\end{tabular}
\end{center}
\caption {Average AUC-ROC its standard deviation for state-of-the-art deep baselines and our proposed approach on the CIFAR-10 and CIFAR-100 datasets.} \label{tab:auc_cifar10_cifar100}
\end{table}

\subsection{Results}

We compare the approaches with respect to their predictive performance, where we report the average area under the receiver operator characteristic curve (AUC-ROC) which is a standard performance metric in anomaly detection~\cite{erfani2016high,Golan2018DeepAD,ruff2018deep}. For the methods that use a non-deterministic algorithm  (CAE/InceptionCAE NN-QED, CAE OCSVM, DeepSVDD, DSEBM), we train 10 models (with different random seeds) and report the average AUC-ROC and its standard deviation over these 10 models for each considered normal class. 


\subsubsection{Results for Q1.\\}

Tables~\ref{tab:auc_mnist_fmnist} and~\ref{tab:auc_cifar10_cifar100} show detailed AUC-ROC scores for state-of-the-art deep baselines and our method. On average, our approach outperforms the deep baselines for all the considered datasets except on Fashion MNIST. Looking at the 50 individual tasks, our InceptionCAE NN-QED method beats DeepSVDD 32 times, DSEBM 48 times, and CAE OCSVM 48 times. The bigger wins come on the more complex CIFAR datasets.

Table~\ref{tab:raw_cae_incpcae} shows how much benefit comes from using our proposed Inception-like CAE architecture with the GAP operation to extract a low-dimensional image representation, compared to using the raw image data or the baseline CAE architecture. On the simpler datasets such as MNIST and Fashion MNIST, both the raw image data and the baseline CAE achieve relatively high AUC-ROCs, but still perform worse than our method. 
However, on the more complex CIFAR-10 and CIFAR-100 datasets, using our more sophisticated approach to learn a low-dimensional representation of the images results in larger improvements in the average AUC-ROCs. 
\begin{table}
\centering
\begin{tabular}{| P{2.5cm} |  P{2.2cm} | P{2.2cm} | P{2.2cm} | }
\cline{1-4}
Dataset              & Raw NN-QED          & CAE NN-QED          & InceptionCAE NN-QED \\ \cline{1-4} 
MNIST               & 94.7 $\pm$ \phantom{1}3.8            & 96.4 $\pm$ \phantom{1}2.4            & 96.9 $\pm$ 1.6 \\ 
Fashion MNIST       & 91.4 $\pm$ \phantom{1}4.9            & 91.6 $\pm$ \phantom{1}4.1            & 93.9 $\pm$ 3.9 \\ 
CIFAR-10            & 59.6 $\pm$ 11.5           & 60.6 $\pm$ 11.6           & 69.6 $\pm$ 4.8 \\ 
CIFAR-100           & 60.2 $\pm$ \phantom{1}9.2            & 62.1 $\pm$ \phantom{1}7.9            & 67.0 $\pm$ 7.1 \\ \cline{1-4}         
\end{tabular}
\caption{Average AUC-ROC and its standard deviation for Raw NN-QED, CAE NN-QED (the baseline CAE architecture), and  InceptionCAE  NN-QED (our proposed approach). The AUC-ROC is averaged over treating each of the ten classes as the normal class.}
\label{tab:raw_cae_incpcae}
\end{table}

\subsubsection{Results for Q2.\\}  

To evaluate the effect of using  product quantization on the predictive performance, we consider computing the exact (squared) Euclidean distance instead of computing the approximate quantized Euclidean distance. 
Again, the quantization is done with the parameters of $m=32$ and $c=4$.
Table \ref{tab:exact-quantized} shows the AUC-ROC for our method using the exact Euclidean distance (variants denoted EED) and quantized Euclidean distance (variants denoted QED) for two representative datasets: Fashion MNIST and CIFAR10. Interestingly, in aggregate using the approximate quantized Euclidean distance slightly improves the predictive performance. Depending on which class is considered normal, there are slight differences in performance between EED and QED: sometimes EED results in a higher AUC-ROC and other times QED does. Using QED to assign the anomaly score is about four times faster than using EED. 
\begin{table}
\centering
\begin{tabular}{| P{2.5cm} | P{2.5cm} | P{2.5cm} | }
\cline{1-3}
Dataset          & InceptionCAE NN-EED     & InceptionCAE NN-QED    \\ \cline{1-3}
Fashion MNIST    & 93.2 $\pm$ 4.1       & 93.9 $\pm$ 3.9       \\ 
CIFAR-10         & 68.3 $\pm$ 5.7       & 69.6 $\pm$ 4.8       \\ \cline{1-3}
\end{tabular}
\caption{The effect of using exact Euclidean distance (variants denoted EED) versus quantized Euclidean distance (variants denoted QED) on predictive performance as measured by AUC-ROC. The AUC-ROC is averaged over treating each of the ten classes as the normal class and the ten models learned for each class. }        \label{tab:exact-quantized}
\end{table} 

\subsubsection{Results for Q3.\\} 

To further investigate where the gains of our approach come from, we explore the effect of the method for assigning an anomaly score on the predictive performance. We compare using the nearest neighbors approach with the quantized Euclidean distance (NN-QED) to using a test image's reconstruction error (RE) as has been done in past work (e.g.,~\cite{sakurada2014anomaly,xia2015learning}). 

Table~\ref{tab:re-v-QED} shows the results on the Fashinon MNIST and CIFAR10 datasets for the baseline CAE and our Inception-like CAE architectures with both methods for computing an anomaly score.  We see that using the distance-based approach results in much better performance than using the reconstruction error. Hence, it is probably worth further exploring using distance-based approaches on top of a bottleneck layer.   

\begin{table}
\begin{center}
    \begin{tabular}{ c | c | c | c |}
\cline{3-4}
 \multicolumn{2}{c|}{}   & RE & NN-QED \\ \cline{2-4} Fashion
    & CAE             & 82.3 $\pm$ 10.9   & 91.6 $\pm$ 4.2\\  MNIST
    & InceptionCAE    & 88.1 $\pm$ \phantom{1}6.5    & 93.9 $\pm$ 3.9  \\  \cline{2-4}

    & CAE             & 56.7 $\pm$ 13.3 & 60.6 $\pm$ 11.6 \\ CIFAR-10
    & InceptionCAE    & 55.3 $\pm$ 14.3 & 69.6 $\pm$ \phantom{1}4.8  \\ \cline{2-4}

    \end{tabular}
\end{center}
\caption {Average AUC-ROC when using the reconstruction error (RE) versus the nearest-neighbors approach with quantized Euclidean distance (NN-QED) for assigning the anomaly score. The AUC-ROC is averaged over treating each of the ten classes as the normal class and the ten models learned for each class. } \label{tab:re-v-QED}
\end{table}

\subsubsection{Results for Q4.\\}\label{sec:quantParams}

To explore how the quantization parameters affect predictive and runtime performance, we try all combinations of parameters $m\in\{1,2,4,8,16,32,64,128 \}$ and $c \in \{1,2,3,4,5,6,7,8\}$. We omit the reduction in memory footprint of using product quantization as the memory tradeoffs are well understood and easily derivable based on the values $m$ and $c$ (see~\cite{DBLP:journals/pami/JegouDS11}). 

\begin{figure}
\centering
\subfloat[Average AUC-ROC.]{%
  \includegraphics[width=0.5\textwidth]{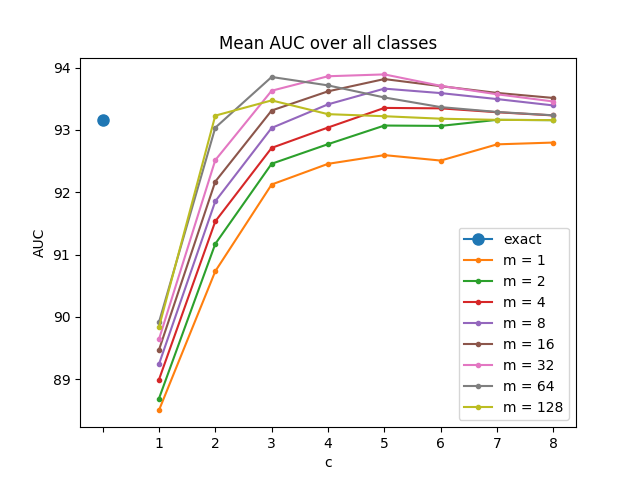}%
  \label{fig:qparams_auc}} 
\subfloat[Prediction time in seconds for the $10,000$ test images.]{%
  \includegraphics[width=0.5\textwidth]{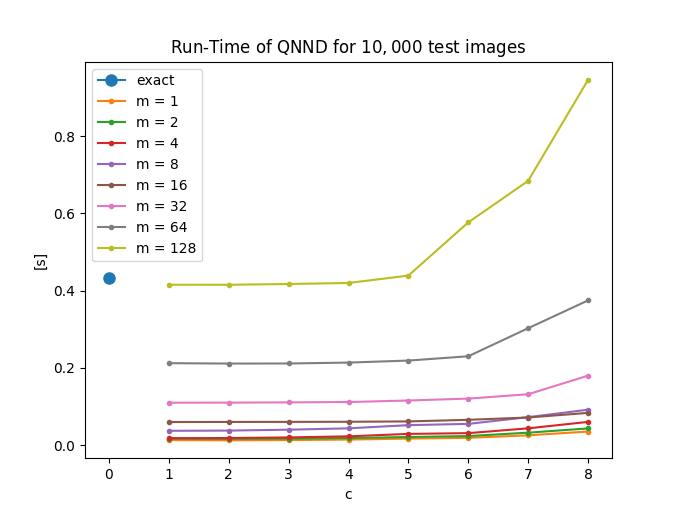}%
   \label{fig:qparams_searchRuntime}}
\caption{Effect of the quantization parameters $m$ and $c$ on  (a) the average AUC-ROC and (b) the prediction time for the test images in seconds for InceptionCAE NN-QED on the CIFAR-10 dataset. The point "exact" represents computing the exact Euclidean distance (i.e., no product quantization).
}
\label{fig:gtd}
\end{figure}

Figure \ref{fig:qparams_auc} shows how the average AUC-ROC (averaged over both treating each class as the normal one and the ten models learned for each class) depends on these parameters for our InceptionCAE NN-QED method on the Fashion MNIST dataset. We see that using values of $c < 3$ has a significant negative effect on the results. The value of $m$ has less of an impact as for a fixed $c$ the average AUC-ROC only varies within a small range regardless of $m$'s value.  Until $m=64$, the AUC-ROC increases with $m$. 

Figure \ref{fig:qparams_searchRuntime} shows how the QED search runtime depends on these parameters. We see that for $m \leq 32$ the QED run-time is significantly smaller than the one for the exact distance search, and for these values of $m$ the QED runtime varies only within a small range with the parameters change. 

\section{Conclusion}
This paper explored anomaly detection in the context of images. We proposed a novel convolutional auto-encoder architecture to learn a low-dimensional representation of the images. Our architecture had two innovations: the use an  Inception-like layer and the application of a GAP operation. Then we assigned an anomaly score to images using a nearest neighbors approach in the learned representation space. Our contribution was to use  product quantization to improve run time performance of this step.  We performed an extensive experimental comparison to both state-of-the-art deep and shallow baselines on four standard datasets. We found that our method resulted in improved predictive performance.

\subsection*{Acknowledgements}
We thank Lukas Ruff from TU Berlin for help reproducing the results from \cite{ruff2018deep}. This research has been partially funded by the European Union’s Horizon  2020 research and innovation program under the Marie Skłodowska-Curie grant agreement No. 752907. JD is partially supported by KU Leuven Research Fund (C14/17/07, C32/17/036), Research Foundation - Flanders (EOS No. 30992574, G0D8819N), VLAIO-SBO grant HYMOP (150033), and the Flanders AI Impulse Program.
%
\bibliographystyle{splncs04}
\bibliography{crc}

\end{document}